\documentclass[runningheads]{llncs}
\usepackage[T1]{fontenc}
\usepackage{newtxtext}       %
\usepackage[varvw]{newtxmath}       

\usepackage{graphicx}
\usepackage{tabularx}
\usepackage{booktabs}
\begin{document}
%
\title{Interaction-Gated Time–Frequency Learning for Parameter-Efficient PPG-based Heart Rate Estimation}
\titlerunning{Interaction-Gated Time–Frequency Learning for PPG HR Estimation}

\author{Anh Khoi Pham\inst{1}\orcidID{0009-0005-9381-4867} \and
Phat Lam\inst{2}\orcidID{0009-0003-5105-5976}
\and 
Thien Truong Nguyen Ly\inst{3}\orcidID{0000-0002-7422-4600}
\and Lam Pham\inst{4}\orcidID{0000-0001-8155-7553}\thanks{Corresponding author}}
\authorrunning{A. K. Pham et al}
%
\institute{International University, Vietnam National University HCMC,
Ho Chi Minh City, Vietnam
\and
University of New Brunswick, Fredericton, Canada
\and Ho Chi Minh City University of Technology, Ho Chi Minh City, Vietnam
\and 
Austrian Institute of Technology, Vienna, Austria
}
\maketitle              
\begin{abstract}
Photoplethysmography (PPG) enables low-cost and non-invasive heart rate (HR) monitoring but remains sensitive to waveform distortion, physiological variability, and measurement interference. Temporal waveform characteristics and frequency-domain periodicity provide complementary information for HR estimation; however, effectively exploring and integrating these representations within a compact model remains challenging. This paper proposes \textit{TF-IG}, a parameter-efficient interaction-gated time–frequency learning framework for PPG-based HR estimation. \textit{TF-IG} employs a dual-branch architecture in which a temporal branch captures pulse morphology via dilated depthwise-separable convolutions, while a frequency branch extracts cardiac periodicity information from bounded log-spectrogram representations. To combine the two domains, we introduce an interaction-gated fusion module that explicitly models feature agreement and discrepancy, while also learning feature-wise weights to balance temporal and spectral information. A parameter-shared regression head further maps the temporal, spectral, and fused representations into the final HR prediction space. Experiments on the benchmark BIDMC PPG and Respiration Dataset using a subject-exclusive training, validation, and test protocol show that \textit{TF-IG} achieves an MAE of $1.227$ BPM, a RMSE of $2.060$ BPM, Pearson correlation of $0.9934$, and $92.28\%$ of estimates within 3 BPM, while remaining parameter-efficient with only 84.6K parameters. Ablation results further demonstrate the complementary contributions of the temporal and frequency branches, interaction-gated fusion, and shared regression head. These results indicate that explicitly modeling cross-domain interactions provides an effective accuracy–complexity trade-off for lightweight PPG-based HR estimation.

\end{abstract}

\section{Introduction}

Continuous heart-rate (HR) monitoring plays an important role in clinical observation, fitness assessment, ambulatory healthcare and long-term physiological monitoring~\cite{ppg_review_01},~\cite{ppg_review_healthcare}.
Although electrocardiography (ECG) remains a standard reference for cardiac monitoring, photoplethysmography (PPG) provides an attractive alternative for continuous and wearable sensing because it measures peripheral blood-volume variations using inexpensive and non-invasive optical hardware~\cite{ppg_review_04},~\cite{review_ppg_05}. Its compact sensing configuration has consequently enabled widespread integration into wearable devices, pulse oximeters, smartphones, and other resource-constrained monitoring platforms. PPG therefore represents a practical sensing modality for scalable and continuous cardiovascular monitoring~\cite{ppg_review_04},~\cite{ppg_review_healthcare}.
Despite these advantages, reliable HR estimation from PPG remains challenging. The measured waveform is influenced not only by cardiac pulsation but also by physiological and acquisition-related factors such as respiration, body site, tissue characteristics, ambient illumination, applied pressure, and artifacts~\cite{intro1}.

A broad range of signal-processing and learning-based approaches has therefore been developed for PPG-based HR estimation. Classical methods typically rely on filtering, peak detection, spectral analysis, or handcrafted signal descriptors~\cite{ppg_review_methods},~\cite{ppg_review_methods_02}. More recently, deep learning has enabled HR-relevant features to be learned directly from PPG observations rather than relying entirely on manually designed processing rules. For example, PPGNet~\cite{ppgnet} proposed an end-to-end network operating directly on short PPG segments for device-independent HR estimation; Deep PPG~\cite{deepppg} leverages convolutional neural networks over spectrogram of PPG and accelerometer signals; Ismail et al.~\cite{recurrent_regressor} further explored a set of features characterizing the signal, modeling by a convolutional-recurrent framework, etc. These studies
demonstrates that temporal and spectral representations of PPG can provide useful information for HR estimation.

The complementarity of temporal and spectral domain follows naturally from the characteristics of PPG. The temporal domain preserves local pulse morphology and short-term waveform dynamics, whereas the frequency domain makes periodic cardiac activity more explicit. Time-frequency representations further characterize how this spectral structure evolves over the observation interval, emphasizing different aspects of the same underlying physiological process~\cite{recurrent_regressor}. 
Although prior work has demonstrated the utility of signal, spectral, and time-frequency representations, effectively structuring the interaction between complementary representations remains an important design problem.
The availability of multiple representations does not by itself determine how they should be combined~\cite{multifusion}. Under signal variability or corruption, temporal and spectral features may provide partly consistent yet partly different evidence about the underlying HR. Therefore, an effective fusion mechanism should explicitly characterize where they agree, where they differ, and how strongly each should jointly contribute to the final representation.

Motivated by these observations, we propose \textit{TF-IG}, a parameter-efficient Interaction-Gated Time--Frequency learning framework for PPG-based HR estimation. First, \textit{TF-IG} learns complementary HR representations through two lightweight branches: a temporal branch based on dilated depthwise-separable convolutions and temporal attention pooling, and a frequency branch that processes a bounded log-spectrogram using spectro-temporal attention pooling. Second, these representations are integrated through a structured interaction-gated mechanism. Element-wise multiplication and absolute difference explicitly encode cross-domain co-activation and discrepancy, while a feature-wise gate adaptively balances the temporal and spectral embeddings. The resulting fused representation, together with the individual branch embeddings, is processed by a shared regression head that provides a common mapping to HR, with the three predictions subsequently combined through learned normalized weights. In this way, \textit{TF-IG} jointly addresses complementary representation learning, explicit cross-domain interaction, and parameter-efficient prediction within a unified lightweight framework.
Accordingly, the main contributions of this work are summarized as follows:

\begin{itemize}
    \item \textbf{Compact dual-domain representation learning.}
    We propose a parameter-efficient dual-branch architecture that jointly captures temporal pulse morphology and spectro-temporal cardiac periodicity using dilated depthwise-separable convolutions and attention-based pooling.

    \item \textbf{Structured cross-domain fusion and shared prediction.}
    We introduce an interaction-gated fusion mechanism that explicitly models temporal--spectral co-activation and discrepancy through element-wise multiplication and absolute difference, while a feature-wise gate adaptively balances the contribution of the two domains. A parameter-shared regression head is further applied to the temporal, frequency, and fused embeddings, providing a common space for HR prediction mapping while avoiding separate prediction networks.

    \item \textbf{Empirical validation of the proposed design.}
    On a subject-independent BIDMC split, \textit{TF-IG} achieves an MAE of $1.227$ BPM, an RMSE of $2.060$ BPM, a Pearson correlation of $0.9934$, and a CP$_3$ of $92.28\%$ using only 84.6K parameters. Ablation experiments further demonstrate the complementary contributions of dual-domain representation learning, interaction-gated fusion, and shared regression to the final performance.
\end{itemize}

\section{Methodology}
This section first formulates PPG-based heart-rate estimation and introduces the notation used throughout the paper Section~\ref{subsec:problem_definition}. Section~\ref{subsec:architecture} describes the overview of the proposed architecture. Subsequently, Section~\ref{subsec:signal_preprocessing} describes the signal preprocessing stage. Individual components within the framework such as time-domain branch, frequency-domain branch, fusion module, and regression head are detailed in Sections~\ref{subsec:time_branch}, \ref{subsec:freq_branch}, \ref{subsec:fusion}, and \ref{subsec:head}, respectively.

\subsection{Problem Definition and Notation}
\label{subsec:problem_definition}


Let $\mathbf{x}_{\mathrm{raw}} \in \mathbb{R}^{T_{\mathrm{raw}}}$ denote a single-channel PPG segment of duration $T_{\mathrm{win}}$ seconds recorded at an original sampling rate $f_s$. Each segment is associated with a reference heart rate $y \in \mathbb{R}$, expressed in beats per minute (BPM). After preprocessing and resampling, the segment is represented as $\mathbf{x} \in \mathbb{R}^{T}$.

The objective is to learn a regression function
$\Phi(\cdot;\Theta)$, parameterized by $\Theta$, that maps the input PPG segment to its corresponding heart-rate estimate:
\begin{equation}
    \hat{y} = \Phi(\mathbf{x};\Theta),
\end{equation}
where $\hat{y} \in \mathbb{R}$ denotes the predicted heart rate. The parameters $\Theta$ are optimized by minimizing a regression loss between $\hat{y}$ and the reference value $y$ over the training set.

Throughout this work, the notation $[\cdot\,;\cdot]$ represents concatenation along the feature dimension, $\odot$ denotes the Hadamard product, and $\sigma(\cdot)$ denotes the sigmoid function.
\begin{figure*}[t]
    \centering
    \includegraphics[width=\linewidth, height = 0.52\linewidth]{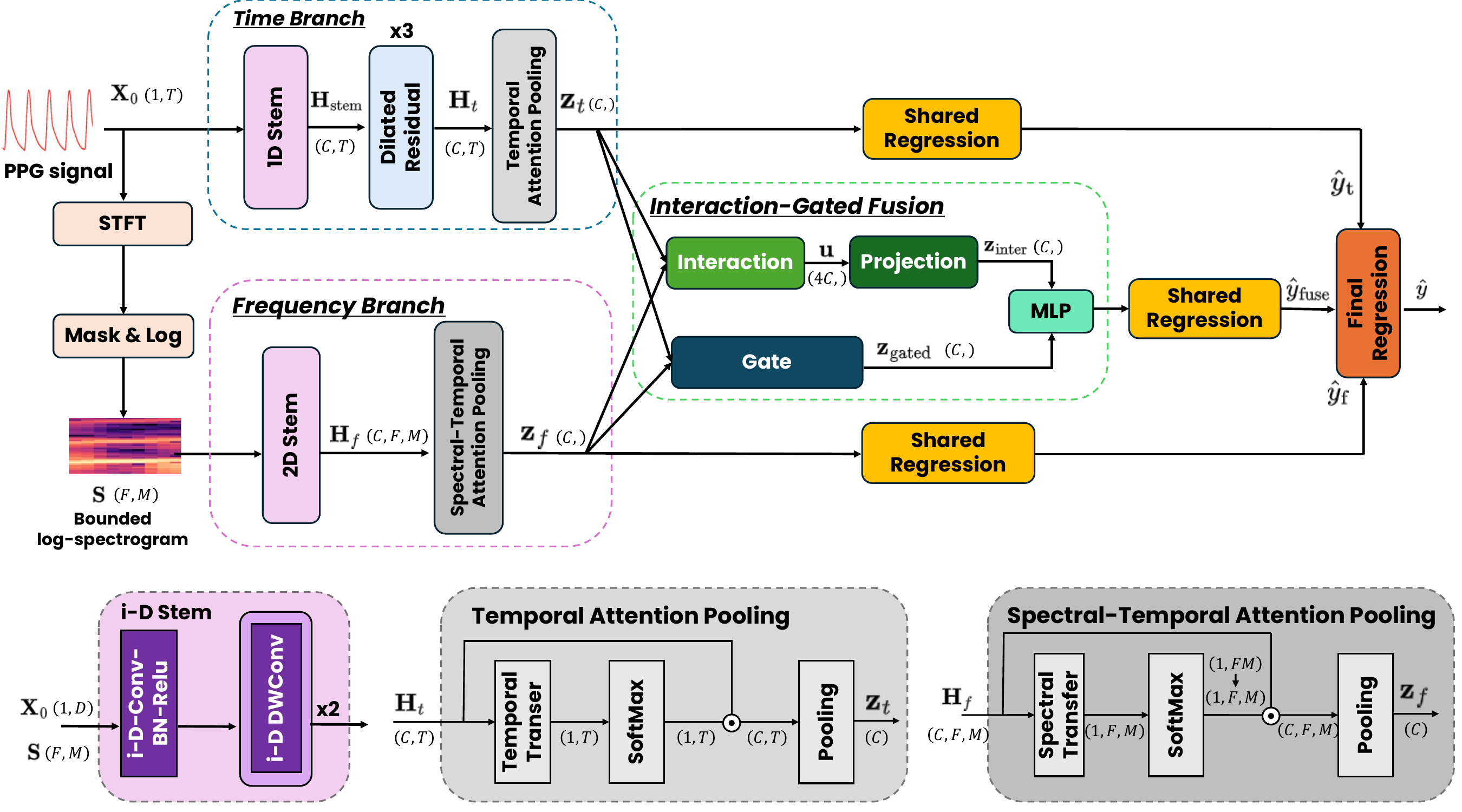}
    \caption{Overview structure of proposed framework}
    \label{fig:model_structure}
\end{figure*}

\subsection{Overall Architecture}
\label{subsec:architecture}


As illustrated in Fig.~\ref{fig:model_structure}, \textit{TF-IG} adopts two complementary processing branches. The time-domain branch operates directly on the PPG waveform and captures local pulse morphology and temporal dependencies. In parallel, the frequency-domain branch transforms the waveform into a bounded log-magnitude spectrogram and models cardiac periodicity over time-frequency transformed representation.

To reduce the computational operations, both streams use depthwise-separable convolutions~\cite{depthwise} as their primary building block. Both branches share a structured processing flow comprising a convolutional stem, domain-specific feature refinement, and targeted attention pooling. Their outputs are converted into fixed-dimensional embeddings through domain-specific attention pooling. An interaction-gated fusion module then combines these embeddings by jointly modeling their agreement, discrepancy, and relative contribution. Finally, a shared regression head generates predictions from the time-domain, frequency-domain, and fused representations, which are combined through a learnable ensemble.

\subsection{Preprocessing}
\label{subsec:signal_preprocessing}

Continuous PPG recordings are first partitioned into fixed-duration windows of $T_{\mathrm{win}}$ seconds using a predefined window stride. Each window is resampled to a target sampling rate $f_s^{*}=62.5$ Hz and then normalized using z-score. The resulting signal $\mathbf{x}\in\mathbb{R}^{T}$ is provided to both the time-domain and frequency-domain branches.

\subsection{Time-Domain Branch}
\label{subsec:time_branch}

\subsubsection{1D Convolutional Stem and Dilated Residual Blocks:}
\label{subsec:time_stem}
The resampled waveform is first arranged as a single-channel sequence,
$\mathbf{X}_0 \in \mathbb{R}^{1\times T}$.  A one-dimensional convolutional stem extracts low-level morphological features while preserving the temporal resolution. The stem consists of a standard Conv1D layer ($K=7$) followed by two dilated depthwise-separable Conv1D layers ($K=5$, with dilations $d=2$ and $d=4$, respectively). Batch normalization and activations are applied after each layer, yielding $\mathbf{H}_{\text{stem}} \in \mathbb{R}^{C \times T}$.

Furthermore, to expand the temporal receptive field without losing resolution, $\mathbf{H}_{\text{stem}}$ is subsequently refined through three cascaded residual blocks. The transformation at the $i$-th block is formulated as:

\begin{equation}
\mathbf{H}_i = \operatorname{ReLU}\big(\mathbf{H}_{i-1} + \Psi_i(\mathbf{H}_{i-1})\big), \quad i \in \{1, 2, 3\},
\label{eq:resblock}
\end{equation}
where $\mathbf{H}_0 = \mathbf{H}_{\text{stem}}$. The function $\Psi_i(\cdot)$ denotes a residual mapping comprising a dilated depthwise-separable Conv1D with kernel size $K_i$ and dilation rate $d_i$, followed by a $1 \times 1$ pointwise projection, batch normalization, and dropout. The output feature map is denoted as $\mathbf{H}_t \triangleq \mathbf{H}_3 \in \mathbb{R}^{C \times T}$.

\subsubsection{Temporal Attention Pooling (TAP):}
\label{subsec:tap}

Temporal attention pooling converts $\mathbf{H}_t$ into a fixed-dimensional embedding while allowing the model to emphasize informative portions of the PPG waveform. Attention logits are generated via a Temporal Transfer block including a two-layer $1\times1$ convolutional network:
\begin{equation}
    \boldsymbol{\ell}_t
    =
    \operatorname{Conv}_{1\times1}^{(2)}
    \left(
        \tanh\left(
            \operatorname{Conv}_{1\times1}^{(1)}(\mathbf{H}_t)
        \right)
    \right),
\end{equation}
where the first projection reduces the channel dimension from $C$ to a hidden dimension $C_h$, and the second produces one scalar logit per temporal position.

The normalized attention weights are
\begin{equation}
    \alpha_k
    =
    \frac{\exp(\ell_{t,k})}
    {\sum_{j=1}^{T}\exp(\ell_{t,j})},
    \qquad k=1,\ldots,T.
    \label{eq:temporal_attention_weights}
\end{equation}
The time-domain embedding $\mathbf{z}_t\in\mathbb{R}^{C}$ is then computed as
\begin{equation}
    \mathbf{z}_t
    =
    \sum_{k=1}^{T}
    \alpha_k\mathbf{H}_t[:,k].
    \label{eq:tap_embedding}
\end{equation}
Unlike uniform global pooling, the TAP operation aims to assign greater importance to waveform intervals containing reliable pulse morphology.

\subsection{Frequency-Domain Branch}
\label{subsec:freq_branch}

\subsubsection{Bounded Log-Spectrogram Representation:}
\label{subsec:spectrogram}
The frequency branch started by transforming the resampled signal $\mathbf{x} \in \mathbb{R}^T$ into spectro-representation $\mathbf{X}_{\mathrm{STFT}}$ via the Short-Time Fourier Transform (STFT) using a Hann window.
Since the dominant cardiac periodicity is expected to lie within a bounded physiological frequency range, we therefore retain only the frequency bins within the cardiac band of $[0.5, 5.0]$ Hz to suppress irrelevant low-frequency drift and high-frequency components and to reduce the dimensionality of the spectral input.

Logarithmic dynamic-range compression is then applied as:
\begin{equation}
    \mathbf{S}[k,m]
    =
    \log\left(
        1+
        \left|\mathbf{X}_{\mathrm{STFT}}[k,m]\right|
    \right),
    \qquad k\in\mathcal{K}.
    \label{eq:log_spectrogram}
\end{equation}

The resulting bounded log-spectrogram is represented as
\begin{equation}
    \mathbf{S}\in\mathbb{R}^{F\times M},
\end{equation}
where $F=|\mathcal{K}|$ is the number of retained frequency bins and $M$ is the number of STFT frames.

\subsubsection{2D Convolutional Stem:}
\label{subsec:freq_stem}
Equivalent to its 1D counterpart, $\mathbf{S}$ is processed by a 2D convolutional stem, which share an identical structural topology. It consists of a standard Conv2D ($K=7$) followed by two dilated depthwise-separable Conv2D layers ($K=5$, dilations $d=2, 4$). Operating over the spatial $F \times M$ plane, this block extracts multi-scale spectro-temporal features, outputting $\mathbf{H}_f \in \mathbb{R}^{C \times F \times M}$.

\subsubsection{Spectro-Temporal Attention Pooling:}
\label{subsec:sta}
As a non-stationary rhythmic signal, heart rate exhibits time-varying spectral energy distributed along instantaneous-frequency trajectories in the time-frequency plane \cite{freq_p1}. Driven by these dynamic properties, Spectro-Temporal Attention Pooling module aggregates the joint spectro-temporal feature map $\mathbf{H}_f$ into a fixed-length embedding $\mathbf{z}_f \in \mathbb{R}^C$ to explicitly capture those joint time-frequency relationships.


Given $\mathbf{H}_f$, a Spectral Transfer block comprising a $1\times1$ convolution, batch normalization, ReLU activation, and a second $1\times1$ convolution produces a logit map
$\mathbf{L}_f\in\mathbb{R}^{F\times M}$. A spatial softmax is applied jointly over all $FM$ positions:
\begin{equation}
    \beta_{u,v}
    =
    \frac{\exp(\mathbf{L}_f[u,v])}
    {\displaystyle
     \sum_{i=1}^{F}\sum_{j=1}^{M}
     \exp(\mathbf{L}_f[i,j])}.
    \label{eq:spectro_temporal_weights}
\end{equation}

The frequency-domain embedding $\mathbf{z}_f\in\mathbb{R}^{C}$ is obtained as
\begin{equation}
    \mathbf{z}_f
    =
    \sum_{u=1}^{F}\sum_{v=1}^{M}
    \beta_{u,v}\mathbf{H}_f[:,u,v].
    \label{eq:sta_embedding}
\end{equation}

\subsection{Interaction-Gated Fusion Module}
\label{subsec:fusion}
While direct concatenation or linear projection can effectively combine time- and frequency-domain representations, it does not explicitly model the cross-domain interactions between these two domains. Therefore, our aim is to achieve an explicit module that can interactively aggregated features from both time and frequency domain.
Specifically, Interaction-Gated Fusion Module decomposes cross-domain fusion into two complementary components: an explicit interaction pathway that models feature relationships and a sample-dependent gate that balances the contribution of the two domains.

\subsubsection{Explicit Cross-Domain Interaction}
\label{subsec:cross_domain_interact}
To capture joint cross-domain relationships, we construct an extended interaction vector $\mathbf{u} \in \mathbb{R}^{4C}$ by combining linear terms, element-wise correlation, and absolute disparity:
\begin{equation}
\mathbf{u} = \left[ \mathbf{z}_t \,;\, \mathbf{z}_f \,;\, \mathbf{z}_t \odot \mathbf{z}_f \,;\, |\mathbf{z}_t - \mathbf{z}_f| \right].
\label{eq:interaction_vec}
\end{equation}
Here, the multiplicative term of $\mathbf{z}_t \odot \mathbf{z}_f$ emphasizes the co-occurring activations across both domains, while the point-wise difference $\vert{}\mathbf{z}_t - \mathbf{z}_f\vert{}$ measures feature discrepancies. The extended vector is subsequently projected back to dimension $C$ via a non-linear projection block:
\begin{equation}
\mathbf{z}_{\text{inter}} = \text{ReLU}\left( \mathbf{W}_u \mathbf{u} + \mathbf{b}_u \right),
\label{eq:interaction_proj}
\end{equation}
where $\mathbf{z}_{\text{inter}} \in \mathbb{R}^C$, $\mathbf{W}_u \in \mathbb{R}^{C \times 4C}$, and $\mathbf{b}_u \in \mathbb{R}^C$. Batch normalization and dropout regularization are applied following the projection.

\subsubsection{Feature-Wise Reliability Gating}
\label{subsec:feature_gating}
In parallel, a feature-wise gate is leveraged to estimate the relative channel-wise weight of each domain directly from the concatenated representations:
\begin{equation}
\mathbf{g} = \sigma\left( \mathbf{W}_g [\mathbf{z}_t \,;\, \mathbf{z}_f] + \mathbf{b}_g \right),
\label{eq:gate_vec}
\end{equation}
where $\sigma(\cdot)$ denotes the sigmoid function, $\mathbf{W}_g \in \mathbb{R}^{C \times 2C}$, and $\mathbf{g} \in (0, 1)^C$. The learned gate $\mathbf{g}$ modulates the combination between time and frequency-domain embeddings along each feature channel. The gated representation is computed as
\begin{equation}
\mathbf{z}_{\text{gated}} = \mathbf{g} \odot \mathbf{z}_t + (\mathbf{1} - \mathbf{g}) \odot \mathbf{z}_f.
\label{eq:gated_combination}
\end{equation}

\subsubsection{Fused Representation}
Finally, the gated features $\mathbf{z}_{\text{gated}}$ and the interaction embedding $\mathbf{z}_{\text{inter}}$ are concatenated and projected to form the unified representation:
\begin{equation}
\mathbf{z}_{\text{fuse}} = \text{ReLU}\left( \mathbf{W}_o [\mathbf{z}_{\text{gated}} \,;\, \mathbf{z}_{\text{inter}}] + \mathbf{b}_o \right),
\label{eq:fused_output}
\end{equation}
where $\mathbf{z}_{\text{fuse}} \in \mathbb{R}^C$, $\mathbf{W}_o \in \mathbb{R}^{C \times 2C}$, and $\mathbf{b}_o \in \mathbb{R}^C$, followed by batch normalization and dropout.

\subsection{Shared Regression Head with Learnable Ensemble}
\label{subsec:head}
The domain-specific features $\mathbf{z}_t$ and $\mathbf{z}_f$, together with the unified interaction-gated representation $\mathbf{z}_{\text{fuse}}$, pass through a regression head to produce the HR estimate. Rather than employing a distinct predictor for each representation, we formulate heart rate regression using a single parameter-shared multi-layer perceptron:

\begin{equation}
\hat{y}_m = f_{\text{reg}}(\mathbf{z}_m), \quad m \in \{t, f, \text{fuse}\},
\label{eq:shared_head}
\end{equation}
where $\mathbf{z}_m \in \mathbb{R}^C$ denotes the input feature embedding, and $\hat{y}_m \in \mathbb{R}$ is the predicted scalar heart rate in BPM. The shared mapping $f_{\text{reg}}(\cdot)$ comprises a linear projection to a bottleneck dimension $C/2$, followed by a $\text{ReLU}$ activation, dropout, and a final linear layer outputting the estimated heart rate.

The final HR estimate is obtained by a learnable weighted combination of the three predictions:
\begin{equation}
\hat{y} = w_{\text{fuse}} \hat{y}_{\text{fuse}} + w_t \hat{y}_t + w_f \hat{y}_f,
\label{eq:ensemble_prediction}
\end{equation}
where the ensemble weights $\mathbf{w} = [w_{\text{fuse}}, w_t, w_f]^\top$ are constrained to sum to 1 via a softmax function applied over learnable logit parameters:
\begin{equation}
w_m = \frac{\exp(p_m)}{\sum_{j \in \{t, f, \text{fuse}\}} \exp(p_j)}.
\end{equation}
The shared head encourages all representations to lie in a consistent HR-predictive space, while the normalized ensemble learns the overall contribution of each prediction.

\section{Experimental Settings}
\subsection{Dataset and Evaluation Metrics}
\label{subsec:data_and_metrics}

The experiments were conducted on the BIDMC PPG and Respiration Dataset (BIDMC), which contains physiological waveforms, including PPG, together with corresponding vital-sign numerics derived from recordings in the MIMIC-II database~\cite{bidmc}. To prevent subject-level information leakage and evaluate generalization, we adopted a subject-exclusive split with a fixed random seed of 36. Specifically, the $53$ subjects in the dataset were partitioned into a training set of $31$ subjects, a validation set of $11$ subjects for hyperparameter tuning and early stopping, and a held-out test set of $11$ subjects for final evaluation.

Model performance was evaluated using six complementary metrics: Mean Absolute Error (MAE), Root Mean Squared Error (RMSE), Mean Squared Error (MSE), Pearson correlation coefficient (\(r\)), and the coverage percentages \(\mathrm{CP}_3\) and \(\mathrm{CP}_5\). The first three metrics quantify the magnitude of HR estimation errors, while \(r\) measures the linear agreement between estimated and reference HR values. In addition, \(\mathrm{CP}_k\) denotes the percentage of predictions whose absolute error is within \(k\) BPM of the reference value.

\subsection{Implementation Details}
\label{subsec:implementation}

\subsubsection{Training Setup and Hyperparameters:}
The proposed model was implemented using the PyTorch framework and trained on an NVIDIA GeForce RTX 4050 GPU. Continuous PPG signals were windowed into $30$-second segments with a $2$-second shift between adjacent frames. All models were trained for $150$ epochs using the Adam optimizer~\cite{adam} with a batch size of $128$, an initial learning rate of $\eta = 0.007$, and a weight decay coefficient of $10^{-4}$. The hidden channel $C$ is set to 64. To facilitate stable convergence, ReduceLROnPlateau scheduler was employed to scale the learning rate by a factor of $0.5$ whenever the validation loss stagnated for $8$ consecutive epochs. To maintain reproducible evaluation protocol across all experiment runs, the global random seed was fixed to $36$. The detail code implementation will be released upon acceptance.

The optimization process was driven by a Mean Squared Error (MSE) loss function defined as:
\begin{equation}
\mathcal{L}_{\text{MSE}} = \frac{1}{N} \sum_{i=1}^{N} (y_i - \hat{y}_i)^2,
\end{equation}
where $y_i$ and $\hat{y}_i$ represent the ground-truth and estimated heart rate values (in BPM) for the $i$-th sample, respectively. Best model checkpointing was dictated by the lowest validation Root Mean Squared Error (RMSE) to evaluate performance on the unseen test set.



\section{Results and Discussion}
\subsection{Main Performance Comparison}
\label{subsec:main_results}

To evaluate the effectiveness of the proposed \textit{TF-IG} architecture, we conducted comparative experiments against both a conventional signal-processing baseline, which estimate heart rate from the dominant spectral component of the PPG signal, refered to as Welch PSD~\cite{welch_psd_hr}, and three learning-based PPG baselines: FCN, FCN\_Residual, and FCN\_DCT~\cite{medvse}. All methods were evaluated using the same preprocessing pipeline and subject-level split of the BIDMC dataset to ensure a fair comparison. Table~\ref{tab:main_performance} summarizes the performance of all methods across six evaluation metrics.

\begin{table*}[t]
\centering
\caption{Performance and complexity comparison on the BIDMC test split under identical experimental protocols.}
\label{tab:main_performance}
\renewcommand{\arraystretch}{1.15}
\setlength{\tabcolsep}{4pt}

\resizebox{\textwidth}{!}{%
\begin{tabular}{@{}lcccccccc@{}}
\toprule
\textbf{Model} 
& \textbf{Params} 
& \textbf{FLOPs/sample}
& \textbf{MAE} (BPM) $\downarrow$
& \textbf{RMSE} (BPM) $\downarrow$
& \textbf{MSE} $\downarrow$
& \textbf{Pearson $r$} $\uparrow$
& \textbf{$CP_5$ (\%)} $\uparrow$
& \textbf{$CP_3$ (\%)} $\uparrow$ \\
\midrule

\textbf{TF-IG (Ours)}
& 84,591
& 13.46 M
&  1.2270
& \textbf{2.0601}
& \textbf{4.2440}
& \textbf{0.9934}
& 95.98
& \textbf{92.28} \\
Welch PSD~\cite{welch_psd_hr}
& 0
& N/A 
& \textbf{1.0823}
& 3.2015
& 10.2494
& 0.9810
& 93.24
& 90.79 \\
FCN\_Residual~\cite{medvse}
& 53,713
& 99.52 M
& 1.5584
& 2.1180
& 4.4860
& 0.9918
& \textbf{97.79}
& 87.45 \\

FCN~\cite{medvse}
& 8,945
& 8.47 M
& 2.9695
& 3.7681
& 14.1986
& 0.9824
& 85.72
& 61.10 \\

FCN\_DCT~\cite{medvse}
& \textbf{3,266}
& \textbf{0.573 M}
& 5.8389
& 8.0721
& 65.1582
& 0.9122
& 57.68
& 36.85 \\

\bottomrule
\end{tabular}%
}
\end{table*}


As shown in Table~\ref{tab:main_performance}, \textit{TF-IG} achieved strong performance across nearly all metrics, with an MAE of $1.2270$,BPM, an RMSE of $2.0601$,BPM, and a Pearson correlation coefficient of $r=0.9934$. Compared with the Welch PSD baseline, \textit{TF-IG} produced a slightly higher MAE but substantially lower RMSE and MSE, indicating fewer large estimation errors and a more concentrated overall error distribution. This is further reflected in the strict accuracy metric, where \textit{TF-IG} achieved $CP_3=92.28\%$, compared with $90.79\%$ for Welch PSD. Moreover, \textit{TF-IG} substantially outperformed all learning-based baselines in terms of MAE, RMSE, correlation, and $CP_3$. In particular, its $CP_3$ exceeded the strongest learning-based baseline, FCN\_Residual by $4.83$ percentage points.

\subsection{Model Parameter Analysis}
\label{subsec:complexity}


As shown in Table~\ref{tab:main_performance}, \textit{TF-IG} contains only $84{,}591$ trainable parameters. Although this is higher than the parameter counts of FCN-DCT, FCN, and FCN-Residual due to its dual-domain encoders, attention modules, and interaction-gated fusion, the complete model remains below $0.1$ million parameters. Together with its superior estimation performance, this compact model size demonstrates a favorable trade-off between predictive accuracy and computational complexity.

\subsection{Ablation Study}
\label{subsec:ablation_study}
To evaluate the contribution of each individual component in the proposed \textit{TF-IG} architecture, we conduct an ablation study across systematically isolated variants. The quantitative results are summarized in Table~\ref{tab:ablation}.

\begin{table*}[t]
\centering
\small
\setlength{\tabcolsep}{3.5pt}
\caption{Ablation study evaluating the contribution of individual architectural components.}
\label{tab:ablation}
\renewcommand{\arraystretch}{1.15}
\resizebox{\textwidth}{!}{%
\begin{tabular}{@{}*{4}{c}*{6}{c}@{}}
\toprule
\textbf{Time Branch} & \textbf{Freq Branch} & \textbf{Int-Gated Fusion} & \textbf{Shared Head} & \textbf{MAE} $\downarrow$ & \textbf{RMSE} $\downarrow$ & \textbf{MSE} $\downarrow$ & \textbf{Pearson $r$} $\uparrow$ & \textbf{$CP_3$ (\%)} $\uparrow$ & \textbf{$CP_5$ (\%)} $\uparrow$ \\
\midrule
\checkmark & $\times$ & $\times$ & \checkmark & 2.1677 & 3.3984 & 11.5492 & 0.9862 & 78.52 & 87.77 \\
$\times$ & \checkmark & $\times$ & \checkmark & 2.347 & 4.5638 & 20.8283 & 0.9713 & 83.31 & 89.02 \\
\checkmark & \checkmark & $\times$ & \checkmark & 1.5878 & 2.464 & 6.0713 & 0.9925 & 87.17 & 92.24 \\
\checkmark & \checkmark & \checkmark & $\times$ & 1.502 & 2.4475 & 5.9904 & 0.9926 & 84.75 & 93.85 \\
\checkmark & \checkmark & \checkmark & \checkmark & \textbf{1.227} & \textbf{2.0601} & \textbf{4.244} & \textbf{0.9934} & \textbf{92.28} & \textbf{95.98} \\
\bottomrule
\end{tabular}%
}
\end{table*}
\begin{table*}[t]
\centering
\small
\setlength{\tabcolsep}{5pt}
\caption{Performance of \textit{TF-IG} with different raw PPG window durations before resampling.}
\label{tab:window_sensitivity}
\renewcommand{\arraystretch}{1.15}
\begin{tabularx}{\textwidth}{@{}l*{6}{>{\centering\arraybackslash}X}@{}}
\toprule
\textbf{Raw Window} & \textbf{MAE} (BPM) $\downarrow$ & \textbf{RMSE} (BPM) $\downarrow$ & \textbf{MSE} $\downarrow$ & \textbf{Pearson $r$} $\uparrow$ & \textbf{$CP_3$ (\%)} $\uparrow$ & \textbf{$CP_5$ (\%)} $\uparrow$ \\
\midrule
$16$ seconds & 1.9331 & 3.2854 & 10.7938 & 0.9856 & 82.56 & 89.58 \\
\textbf{$30$ seconds} & \textbf{1.2270} & \textbf{2.0601} & \textbf{4.2440} & \textbf{0.9934} & \textbf{92.28} & \textbf{95.98} \\
\bottomrule
\end{tabularx}
\end{table*}

The single-branch variants demonstrate that both temporal and spectral representations can independently support HR estimation, although neither is sufficient to achieve the performance of the complete model. The time-only variant obtains lower overall errors than the frequency-only variant, with an MAE of $2.1677$~BPM and an RMSE of $3.3984$~BPM, compared with $2.3470$~BPM and $4.5638$~BPM, respectively. In contrast, the frequency-only variant achieves higher $CP_3$ and $CP_5$ values, suggesting that the two branches capture different but complementary characteristics of the PPG signal.

Combining the time and frequency branches without the interaction-gated fusion module substantially improves performance over both single-branch configurations, reducing the MAE to $1.5878$~BPM and increasing the Pearson correlation to $0.9925$. This result confirms that jointly exploiting waveform morphology and spectral periodicity provides a more informative representation than relying on either domain alone. Incorporating the proposed fusion module further reduces the MAE and RMSE to $1.2270$~BPM and $2.0601$~BPM, respectively, while increasing $CP_3$ from $87.17\%$ to $92.28\%$. These improvements indicate that explicitly modeling cross-domain interactions and adaptively balancing the two representations significantly improve model performance. 

The shared regression head also contributes to the final performance. Replacing it with three distinct prediction heads increases the MAE from $1.2270$ to $1.5020$~BPM and decreases $CP_3$ from $92.28\%$ to $84.75\%$. Overall, the full \textit{TF-IG} configuration achieves the best result across all reported metrics, validating the complementary contributions of the dual-branch design, interaction-gated fusion, and parameter-shared regression head.

\subsection{Window Size Sensitivity Analysis}
\label{subsec:window_ablation}
To investigate the effect of the observation duration on HR estimation, we evaluated \textit{TF-IG} using raw PPG windows of $16$ and $30$ seconds. Here, the window size refers to the duration of each raw signal segment before resampling; after segmentation, both configurations were processed using the same resampling pipeline and experimental protocol described above. The results are reported in Table~\ref{tab:window_sensitivity}.


Increasing the raw window duration from $16$ to $30$ seconds consistently improves performance across all evaluation metrics. In particular, the MAE decreases from $1.9331$ to $1.2270$~BPM, while the RMSE decreases from $3.2854$ to $2.0601$~BPM. The longer window also increases $CP_3$ and $CP_5$ by $9.72$ and $6.40$ percentage points, respectively, and raises the Pearson correlation from $0.9856$ to $0.9934$.

These results suggest that the $30$-second raw window provides the model with a broader temporal context and a larger number of cardiac cycles, supporting more stable extraction of waveform morphology and spectral periodicity after resampling. Consequently, the $30$-second configuration is adopted for the main experiments. Nevertheless, the $16$-second setting remains a potential alternative when a shorter observation interval is preferred, albeit with reduced estimation accuracy.



\subsection{Feature Space Representation Analysis}
\label{sec:feature_space_analysis}
\begin{figure*}[t]
    \centering
    \includegraphics[width=\textwidth]{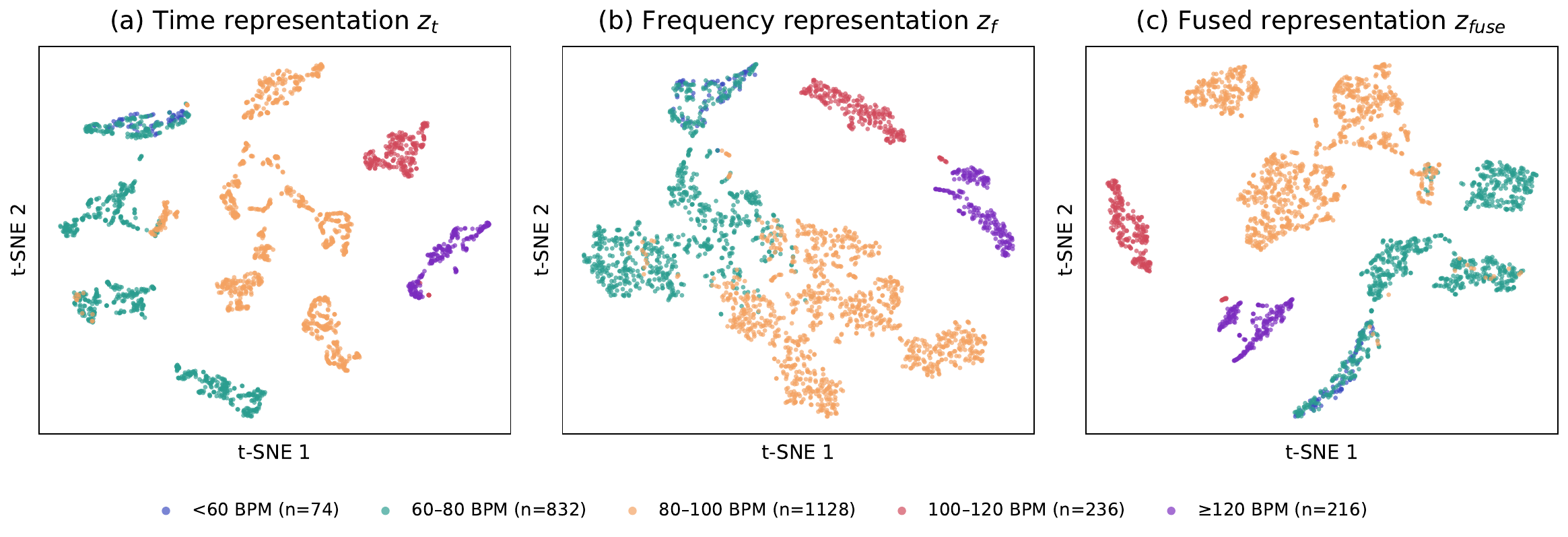}
    \caption{t-SNE visualization of the representations learned by TF-IG on
    the BIDMC test split: (a) temporal features $\mathbf{z}_t$, (b) frequency
    features $\mathbf{z}_f$, and (c) fused features
    $\mathbf{z}_{\mathrm{fuse}}$. Colors indicate reference-HR ranges.}
    \label{fig:feature_space_tsne}
\end{figure*}
To investigate how the two modality-specific branches and the proposed fusion
module organize heart-rate information, we extracted the temporal feature
$\mathbf{z}_t$, frequency feature $\mathbf{z}_f$, and fused feature
$\mathbf{z}_{\mathrm{fuse}}$ for all windows in the held-out BIDMC test
set. Each representation is a $C$-dimensional vector obtained before the
regression head. For qualitative visualization, the features were standardized
independently and projected to two dimensions using t-SNE. Points were colored according to five reference-HR
ranges: $<60$, $60$--$80$, $80$--$100$, $100$--$120$, and $\geq120$ BPM with their corresponding number of windows belong to that range.

The temporal representation is distributed across several separated groups, with samples from the same HR range appearing in different regions. This suggests that the temporal branch can capture not only HR-related dynamics but also variations in waveform morphology, signal quality, and subject-specific characteristics. In contrast, the frequency representation exhibits more compact and localized organization according to HR, consistent with its direct encoding of cardiac periodicity. However, overlap between neighboring HR ranges remains. It may be due to their continuous boundaries and the effects of spectral noise, harmonics, and physiological variability between subjects. The fused representation integrates temporal morphology and spectral periodicity, reducing ambiguities present in either individual branch and producing a more balanced organization of HR-related information. Overall, the visualization supports the complementary roles of the temporal and frequency branches and the effectiveness of their fusion.

\section{Conclusion}

This work introduced \textit{TF-IG}, a parameter-efficient framework for PPG-based heart-rate estimation that jointly models temporal morphology and spectral periodicity. Its interaction-gated fusion captures cross-domain agreement and discrepancy while adaptively balancing their contributions, and a shared regression head provides a compact common mapping to the final HR prediction. Evaluation on the subject-exclusive BIDMC split demonstrates that \textit{TF-IG} consistently outperforms the evaluated baselines across most of error-based, correlation, and strict accuracy metrics, while maintaining a compact parameter footprint. Ablation studies further validate the complementary roles of dual-domain representation learning, interaction-gated fusion, and shared regression in improving estimation performance. Overall, these results demonstrate that explicitly modeling temporal–spectral interactions provides an effective accuracy–complexity trade-off for efficient PPG-based HR estimation.

%
%
%
\bibliographystyle{splncs04}
\begingroup
\raggedright
\bibliography{refs}
\endgroup
\end{document}